# Molecular Structure Extraction From Documents Using Deep Learning

**Joshua Staker[†*], Kyle Marshall[†*], Robert Abel[‡], Carolyn McQuaw[†]**

[†]Schrödinger, Inc., 101 SW Main Street, Portland Oregon 97204, United States
[‡]Schrödinger, Inc., 120 West 45th Street, New York, New York 10036, United States
* {joshua.staker,kyle.marshall}@schrodinger.com

## ABSTRACT

Chemical structure extraction from documents remains a hard problem due to both false positive identification of structures during segmentation and errors in the predicted structures. Current approaches rely on handcrafted rules and subroutines that perform reasonably well generally, but still routinely encounter situations where recognition rates are not yet satisfactory and systematic improvement is challenging. Complications impacting performance of current approaches include the diversity in visual styles used by various software to render structures, the frequent use of ad hoc annotations, and other challenges related to image quality, including resolution and noise. We here present end-to-end deep learning solutions for both segmenting molecular structures from documents and for predicting chemical structures from these segmented images. This deep learning-based approach does not require any handcrafted features, is learned directly from data, and is robust against variations in image quality and style. Using the deep-learning approach described herein we show that it is possible to perform well on both segmentation and prediction of low resolution images containing moderately sized molecules found in journal articles and patents.

## Introduction

For drug discovery projects to be successful, it is often crucial that newly available data are quickly processed and assimilated through high quality curation. Furthermore, an important initial step in developing a new therapeutic includes the collection, analysis, and utilization of previously published experimental data. This is particularly true for small-molecule drug discovery where collections of experimentally tested molecules are used in virtual screening programs, quantitative structure activity/property relationship (QSAR/QSPR) analyses, or validation of physics-based modeling approaches. Due to the difficulty and expense of generating large quantities of experimental data, many drug discovery projects are forced to rely on a relatively small pool of in-house experimental data, which in turn may result in data volume as a limiting factor in improving in-house QSAR/QSPR models.

One promising solution to the widespread lack of appropriate training set data in drug discovery is the amount of data currently being published.[1] Medline reports more than 2000+ new life science papers published per day,[2] and this estimate does not include other literature indexes or patents that further add to the volume of newly published data. Given this high rate at which new experimental data is entering the public literature, it is increasingly important to address issues related to data extraction and curation, and to automate these processes to the greatest extent possible. One such area of data curation in life sciences that continues to be difficult and time consuming is the extraction of chemical structures from publicly available sources such as journal articles and patent filings.

Most publications containing data related to small molecules do not provide the molecular structures in a computer readable format (e.g., SMILES, connection table, etc.). Instead, computer programs are used by authors to draw the corresponding structures, and are included in the document via an image of the resulting drawing. Publishing documents with only images of structures necessitates the manual redrawing of the structures in chemical sketching software as a means of converting the structures into computer readable formats for use in downstream computation and analysis. Redrawing chemical structures can be time consuming and often requires domain knowledge to adequately resolve ambiguities, interpret variations in style, and decide how annotations should be included or ignored.

Solutions for automatic structure recognition have been described previously.[3-9] These methods utilize sophisticated rules that perform well in many situations, but can experience degradation in output quality under commonly encountered conditions, especially when input resolution is low or image quality is poor. One of the challenges to improving current extraction rates is that rule-based systems are necessarily highly interdependent and complex, making further improvements difficult. Furthermore, rule-based approaches can be demanding to build and maintain because they require significant domain expertise and require contributors to anticipate and codify rules for all potential scenarios the system might encounter. Developing hand-coded rules is particularly difficult in chemical structure extraction where a wide variety of styles and annotations are used, and input quality is not always consistent. The goal of this work is twofold: 1) demonstrate it is possible to develop an extraction method to go from input document to SMILES without requiring the implementation of hand-coded rules or features; and 2) further demonstrate it is possible to improve prediction accuracy on low quality images using such a system.

Deep learning and other data-driven technologies are becoming increasingly widespread in life sciences, particularly in drug discovery and development.[10-13] In this work we leverage recent advances in image processing, sequence generation, and computing over latent representations of chemical structures to predict SMILES for molecular structure images. The method reported here takes an image or PDF and performs segmentation using a convolutional neural network. SMILES are then generated using a convolutional neural network in combination with a recurrent neural network (encoder-decoder) in an end-to-end fashion (meaning, the architecture computes SMILES directly from raw images). We report results based on preliminary findings using our deep learning-based method and provide suggestions for potential improvements in future iterations. Using a downsampled version of a published dataset, we show that our deep learning method performs well under low quality conditions, and may operate on raw image data without hand-coded rules or features.

## Related Work

Automatic chemical structure extraction is not a new idea. Park et al.,[3] McDaniel et al.,[5] Sadawi et al.,[6] Valko & Johnson,[7] and Filippov & Nicklaus[8] each utilize various combinations of image processing techniques, optical character recognition, and hand-coded rules to identify lines and characters in a page, then assemble these components into molecular connection tables. Similarly, Frasconi et al.[9] utilize low level image processing techniques to identify molecular components but rely on Markov logic to assemble the components into complete structures. Park et al.[4] demonstrated the benefits of ensembling several recognition systems together in a single framework for improved recognition rates. Currently available methods rely on low-level image processing techniques (edge detectors, vectorization, etc.) in combination with subcomponent recognition (character and bond detection) and high-level rules that arrange recognized components into their corresponding structures.

There are continuing challenges, however, that limit the usefulness of currently available methods. As discussed in Valko & Johnson[7] there are many situations in the literature where designing specific rules to handle inputs becomes quite challenging. Some of these include wavy bonds, lines that overlap (e.g., bridges), and ambiguous atom labels. Apart from complex, ambiguous, or uncommon representations, there are other challenges that currently impact performance, including low resolution or noisy images. Currently available solutions require relatively high resolution input, e.g., 300+ dpi,[5,19] and tolerate only small amounts of noise. Furthermore, rule-based systems can be difficult to improve due to the complexity and interconnectedness of the various recognition components. Changing a heuristic in one area of the algorithm can impact and require adjustment in another area, making it difficult to improve components to fit new data while simultaneously maintaining or improving generalizability of the overall system.

Apart from accuracy of structure prediction, filtering of false positives during the structure extraction process also remains problematic. Current solutions rely on users to manually filter out results, choosing which predicted structures should be ignored if tables, figures, etc. are predicted to be molecules. In order to both improve extraction and prediction of structures in a wide variety of source materials, particularly with noisy or low quality input images, it is important to explore alternatives to rule-based systems. The deep learning-based method outlined herein provides (i) improved accuracy for poor quality images and (ii) a built-in mechanism for further improvement through the addition of new training data.

## Deep Learning Method

The deep learning model architectures for segmentation and structure prediction described in this work are depicted in Figure 1. The segmentation model identifies and extracts chemical structure images from input documents, and the structure prediction model generates a computer-readable SMILES string for each extracted chemical structure image.

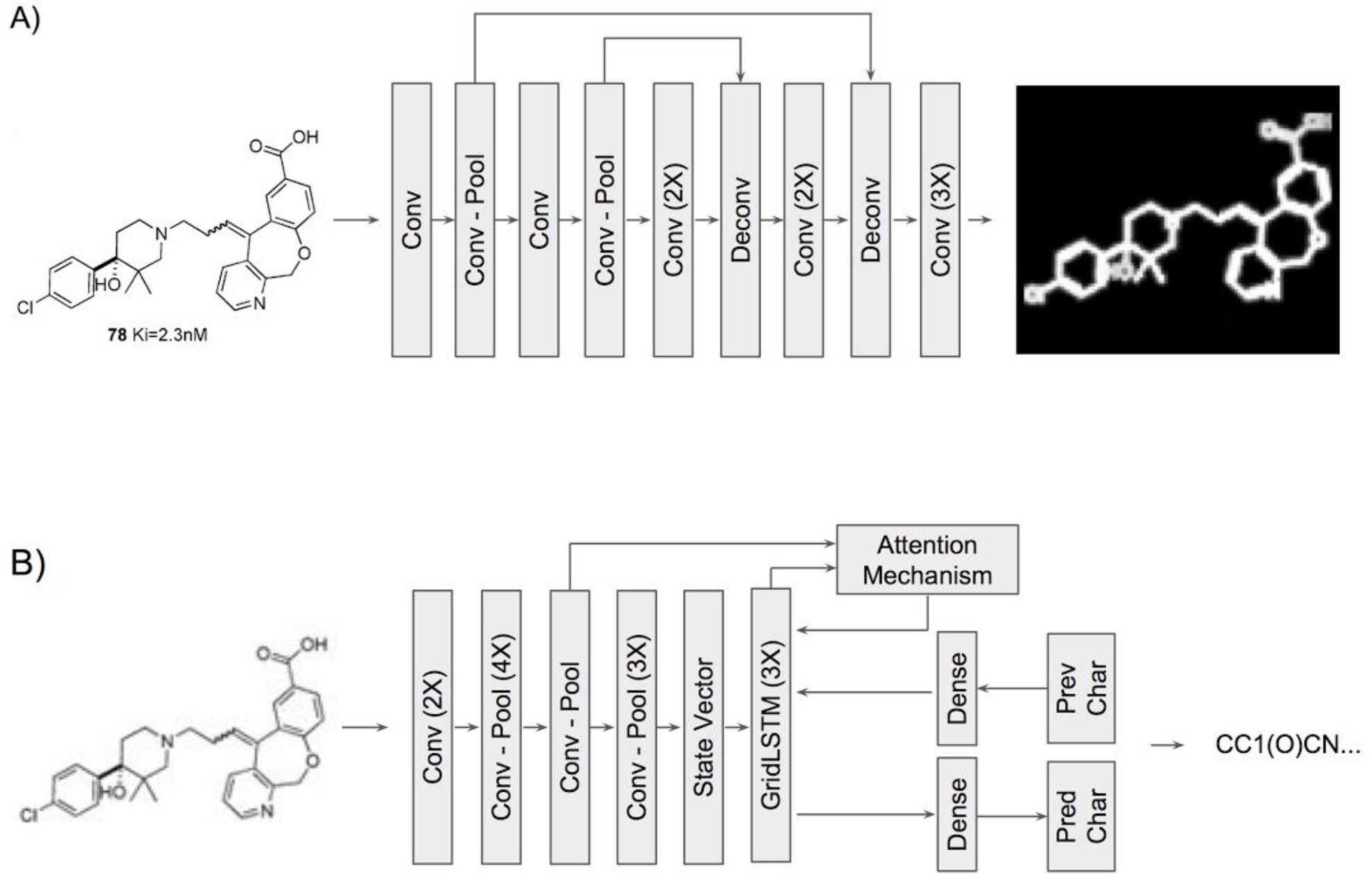

**Figure 1.** In subpanel (A) we depict the segmentation model architecture and in subpanel (B) we depict the structure prediction model architecture. For brevity, similar layers are condensed with a multiplier indicating how many layers are chained together, e.g., “2X”. All convolution (conv) layers are followed by a parametric ReLU activation function. “Pred Char” represents the character predicted at a particular step and “Prev Char” stands for the predicted character at the previous step. Computation flow through the diagrams is left to right (and top to bottom in (B)).

**Segmentation**
When presented with a document containing chemical structures, the initial step in the extraction pipeline is first identifying what are structures and segment these from the rest of the input. Successful segmentation is important to (i) provide cleanest possible input for accurate sequence prediction, and (ii) exclude objects that are not structures but contain similar features, e.g., charts, graphs, logos, and annotations. Segmentation herein utilizes a deep convolutional neural network to predict which pixels in input images are likely to be part of a chemical structure. In designing the segmentation model we followed the "U-Net" design strategy outlined in Ronneberger et al.[20] which is especially well suited for full-resolution detection at the top of the network and enables fine-grained segmentation of structures in the experiments reported here. The U-Net supports full-resolution segmentation by convolving (with pooling) the input to obtain a latent representation, then upsampling the latent representation using deconvolution with skip-connections until the output is at a resolution that matches that of the input. In our experiments, the inputs to our model were preprocessed to be grayscale and downsampled to approximately 60 dpi[1] resolution. We found 60 dpi input resolution to be sufficient for segmentation while providing significant speed improvements versus higher resolutions. We fed the preprocessed inputs into our implementation of the U-Net and the logits generated at the top of the network were scaled using a softmax activation, and provided a predicted probability between 0-1.0 for each pixel, identifying the likelihood pixels belonged to a structure.

The predicted pixels formed masks generated at the same resolution as the original input images and allowed for sufficiently fine-grained extraction. The masks obtained from the segmentation model were binarized to remove low confidence pixels, and contiguous areas of pixels that were not large enough to contain a structure were removed. To remove areas too small to be structures, we counted the number of pixels in a contiguous area and deemed the area a non-structure if the number of pixels was below a threshold. We also tested the removal of long, straight horizontal and vertical lines in the input image using the Hough transform.[21] Line removal improved mask quality in many cases, especially in tables where structures were very close to grid lines, and was included in the final model. Individual entities (a single, contiguous group of positively predicted pixels) in the refined masks were assumed to contain single structures and were used to crop structures from the original inputs, resulting in a collection of individual structure images.

During inference we observed qualitatively better masks when generating several masks at different resolutions and averaging the masks together into a final mask used to crop out structures. Averaged masks were obtained by scaling inputs to each resolution within the range 30 to 60 dpi in increments of 3 dpi, then generating masks for each image using the segmentation model. The resulting masks were scaled to the same resolution (60 dpi) and averaged together. The averaged masks were then scaled to the original input resolution (usually 300 dpi) and then used to crop out individual structures. Figure 2 shows an example journal article page along with its predicted mask.

**Structure Prediction**

The images of individual structures obtained using the segmentation model were automatically transcribed into the corresponding SMILES sequences representing the contained structures using another deep neural network. The purpose of this network was to take an image of a single structure and, in an end-to-end fashion, predict the corresponding SMILES string of the structure contained in the image. The network comprised an encoder-decoder strategy where structure images were first encoded into a fixed-length latent space (state vector) using a convolutional neural network and then decoded into a sequence of characters using a recurrent neural network. The convolutional network consisted of alternating layers of 5x5 convolutions, 2x2 max-pooling, and a parameterized ReLU activation function,[30] with the overall network conceptually similar to the design outlined in Krizhevsky et al.[22] but without the final classification layer. To help mitigate issues in which small but important features are lost during encoding, our network architecture did not utilize any pooling method in the first few layers of the network.

The state vector obtained from the convolutional encoder is passed into a decoder to generate a SMILES sequence. The decoder consisted of an input projection layer, three layers of GridLSTM cells,[23] an attention mechanism (implemented similarly to the soft attention method described in Xu et al.[24] and Bahdanau et al.,[25] and the global attention method in Luong et al.[15]), and an output projection layer. The state vector from the encoder was used to initialize the GridLSTM cell states and the SMILES sequence was then generated a character at a time, similar to the decoding method described in Sutskever et al.[26] (wherein sentences were generated a word at a time while translating English to French). Decoding is started by projecting a special start token into the GridLSTM (initialized by the encoder and conditioned on an initial context vector as computed by the attention mechanism), processing this input in the cell, and predicting the first character of the output sequence. Subsequent characters are produced similarly, with each prediction conditioned on the previous cell state, the current attention, and the previous output projected back into the network. The logits vector for each character produced by the network is of length N, where N is the number of available characters (65 characters in this case). A softmax activation is applied to the logits to compute a probability distribution over characters, and the highest scoring character is selected for a particular step in the sequence. Sequences were generated until a special end-of-sequence token was predicted, at which point the completed SMILES string was returned. During inference we found accuracy improved when predicting images at several different (low) resolutions and returning sequences of the highest confidence, which was determined by multiplying together the softmax output of each predicted character in the sequence.

[1] Resolution is expressed in dpi rather than in raw pixel lengths because the data was derived from PDFs where conversion of the PDF pages into images was a necessary step and was performed in dpi.

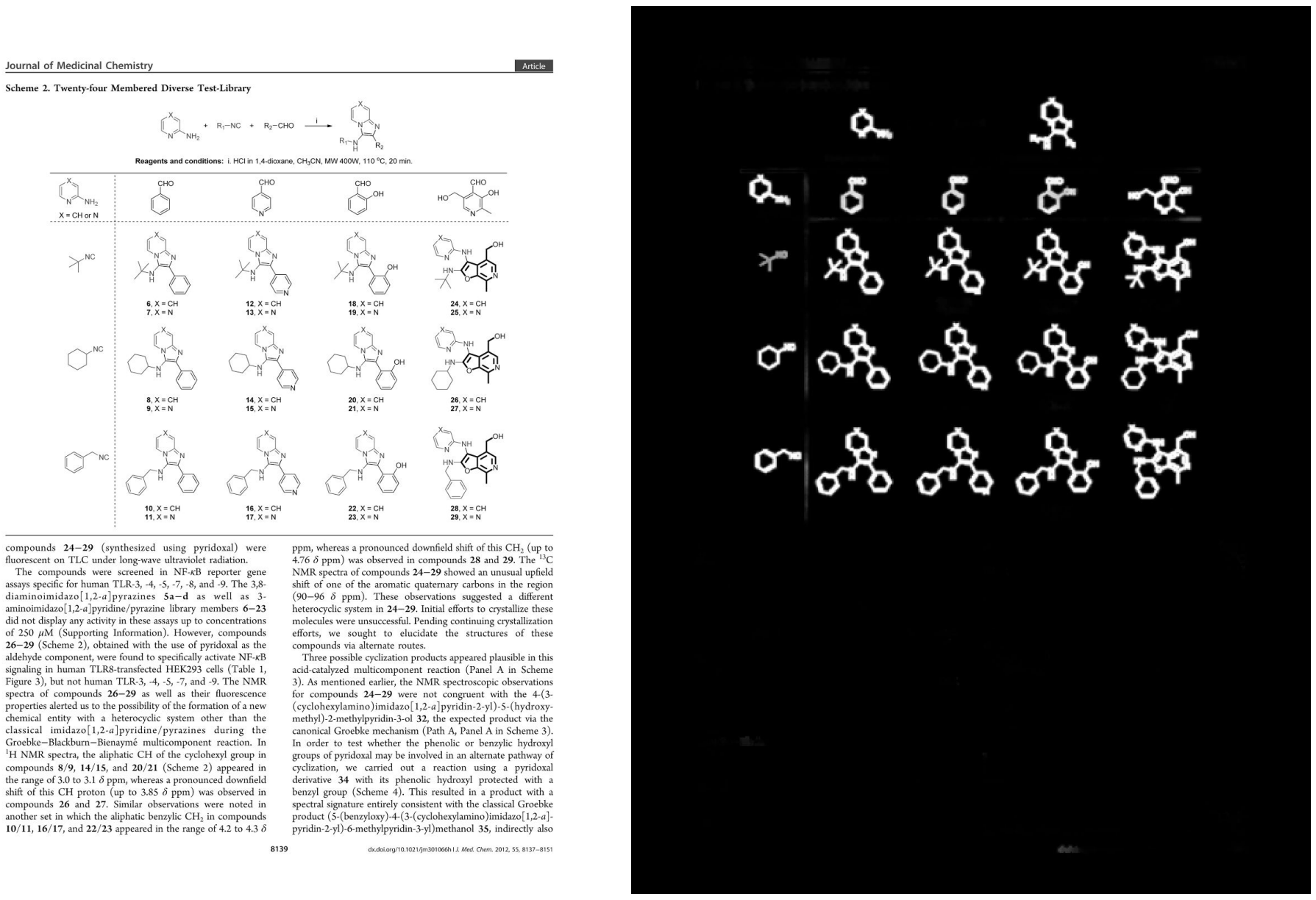

**Figure 2.** An example showing the output of the segmentation model when processing a journal article page from Salunke et al.[14] All of the text and other extraneous items are completely removed with the exception of a few faint lines that are amenable to automated post processing.

The addition of an attention mechanism in the decoder helped solve several challenges. Most importantly, attention enabled the decoder to access information produced earlier in the encoder and minimized the loss of important details that may otherwise be overly compressed when encoding the state vector. Additionally, attention enabled the decoder to reference information closer to the raw input during the prediction of each character and was important considering the significance of pixelwise features in low resolution structure images. See Figure 3 for an example of the computed attention and how the output corresponds to various characters recognized during the decoding process. Apart from using attention for improved performance, the attention output is useful for repositioning a predicted structure into an orientation that better matches the original input image. This is done by converting the SMILES into a connection table using the open source Indigo toolkit,[27] and repositioning each atom in 2D space according to the coordinates of each character's computed attention. The repositioned structure then more closely matches the original positioning and orientation in the input image, enabling users to more easily identify and correct mistakes when comparing the output with the original source.

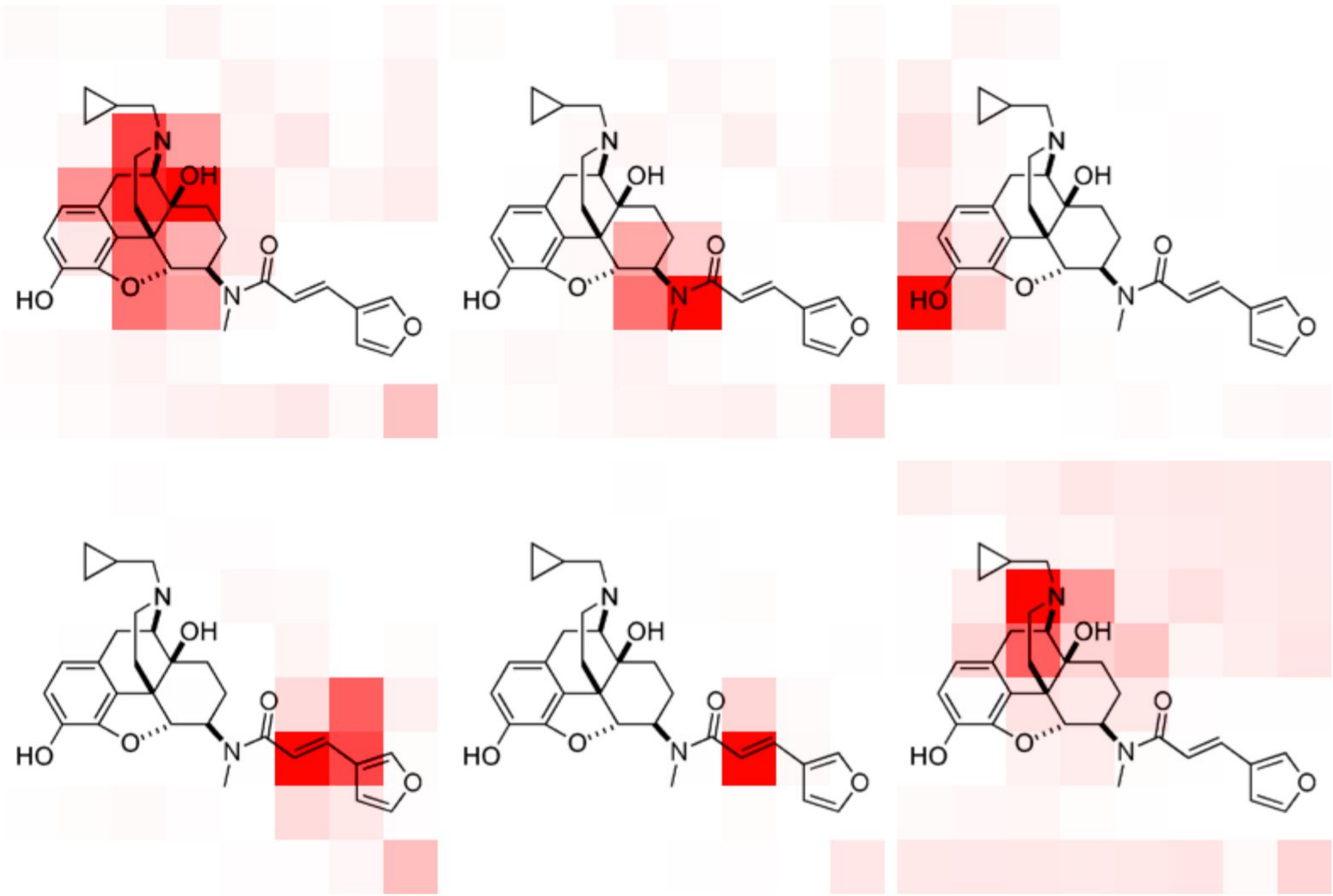

**Figure 3.** Heatmaps are here depicted representing the computed attention during character prediction in the order left to right, top to bottom: [, @, O, /, =, N.

The complete encoder-decoder framework is fully differentiable and was trained end-to-end using a suitable form of backpropagation, enabling SMILES to be fully generated using only raw images as input. During decoding SMILES were generated a character at a time, from left to right. Additionally, no external dictionary was used for chemical abbreviations (superatoms) rather these were learned as part of the model, thus images may contain superatoms and the SMILES are still generated a character at a time. This model operates on raw images and directly generates chemically valid SMILES with no explicit subcomponent recognition required.

## Datasets

**Segmentation Dataset**

To our knowledge, no dataset addressing molecular structure segmentation has been published. To provide sufficient data to train a neural network while minimizing manual effort required to curate such a dataset, we developed a pipeline for automatically generating segmentation data. To programmatically generate data, in summary, the following steps were performed: i) remove structures from journal and patent pages, ii) overlay structures onto the pages, iii) produce a ground truth mask identifying the overlaid structures, and iv) randomly crop images from the pages containing structures and the corresponding mask. In detail, OSRA[8] was utilized to identify bounding boxes of candidate molecules within the pages of a large number of publications, both published journal articles and patents. The regions expected to contain molecules were whited-out, thus leaving pages without molecules. OSRA was not always correct in finding structures and occasionally non-structures (e.g., charts) were removed suggesting that cleaner input may further improve model performance. Next, images of molecules made publically available by the United States Patent and Trademark Office (USPTO)[28] were randomly overlaid onto the pages while ensuring that no structures overlapped with any non-white pixels. Structure images were occasionally perturbed using affine transformations, changes in background shade, and/or lines added around the structure (to simulate table grid lines). We also generated the true mask for each overlaid page; these masks were zero-valued except where pixels were part of a molecule (pixels assigned a value of 1). During training, samples of 128x128 pixels were randomly cropped from the overlaid pages and masks, and arranged into mini-batches for feeding into the network for training; example image-mask pairs are shown in Figure 4.

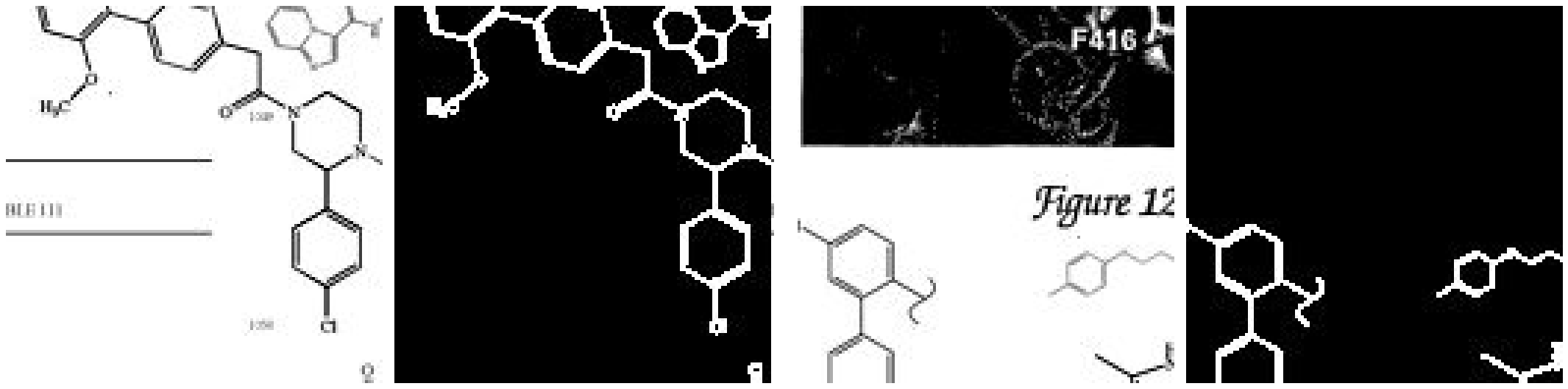

**Figure 4.** Examples sampled from the generated segmentation dataset. Inputs to the network are shown on the left with the corresponding masks used in training shown on the right. White indicates which pixels are part of a chemical structure.

**Molecular Image Dataset**

An important goal of this work was to improve recognition of low resolution or poor quality images. Utilizing training data that is of too high quality or too clean could negatively impact the generalizability of the final model. Arguably, the network should be capable of becoming invariant to image quality when trained explicitly on both high and low quality examples, at the expense of more computation and likely more data. However, we opted to handle quality implicitly by scaling all inputs down considerably. To illustrate the impact of scaling images of molecular structures, consider two structures in Figure 5 that are chemically identical but presented with different levels of quality. The top-left image is of fairly high quality apart from some perforation throughout the image. In the top-right image the perforation is much more pronounced with some bonds no longer continuous (small breaks due to the excessive noise). When these images are downsampled significantly using bilinear interpolation, the images appear similar and it is hard to differentiate which of the two began as a lower quality image, apart from one being darker than the other.

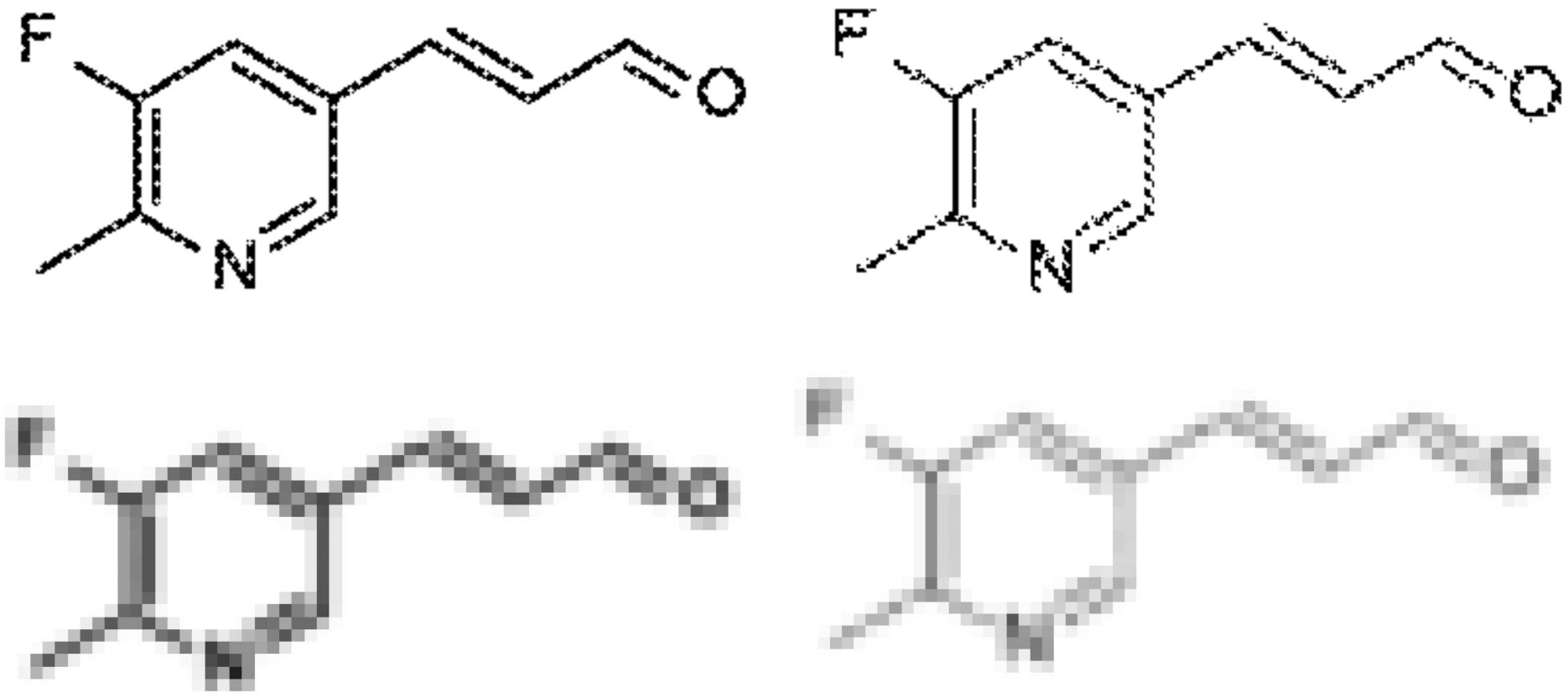

**Figure 5.** Noisy images of the same structure from Bonanomi et al.[16] The top two images are the original images with the corresponding downsampled image below. Image quality appears similar when downsampled considerably.

Chemical structures vary significantly in size, some being small fragments with just a few atoms, to very large structures, including natural products or peptide sequences. This necessitates using an image size that is not too large to be too computationally intensive to train, but large enough to fully fit reasonably sized structures, i.e., drug-like small molecules, into the image. Although structures themselves can be large the individual atoms and their relative connectivity may be contained within a small number of pixels regardless of the size of the overall structure. Thus, scaling an image too aggressively will result in important information being lost. To train a neural network that can work with both low and high quality images, we utilized an image size of 256x256 and scaled images to fit within this size constraint (bond lengths resulting in approximately the 3-12 pixels range). Training a neural network over higher resolution images is an interesting research direction and may improve results, but was here left for future work.

To ensure that the training data contained a variety of molecular image styles we used three separate datasets, each sampled uniformly during training. Additionally, we focused on drug-like molecules and imposed the following restrictions while preparing data and training the model:

- Structures with a SMILES length of 21-100 characters (a range that covers most drug-like small molecules) were included, all others removed.
- Attachment placeholders of the format R1, R2, R3, etc., were included but other forms of placeholder notation were not included.
- Enumeration fragments were not predicted or attached to the parent structure.
- Salts were removed and each image was assumed to contain only one structure.
- All SMILES were kekulized and canonicalized.

The first utilized dataset consisted of a 57 million molecule subset of molecules available in the PubChem database[29] rendered into images of 256x256 pixels of various styles (bond thicknesses, character sizes, etc.) using Indigo. PubChem structures were available in InChI format and were converted to SMILES using Indigo. To evaluate performance of the model during training the dataset was split into train/validation subsets; 90% of the dataset was used to train the model and the remaining 10% was reserved for validation.

The second dataset comprised 10 million images rendered using Indigo in OS X. Because Indigo rendering output can vary significantly between operating systems, we included these images to supplement training with additional image styles.

The third dataset consisted of 1.7 million image/molecule pairs curated from data made publicly available by the USPTO.[28] Many of these images contain extraneous text and labels and were preprocessed to remove non-structure elements before training. For some files in the USPTO dataset, we observed that Indigo does not correctly retain stereochemistry when converting MOL format into canonical SMILES, which resulted in some SMILES not containing identical stereochemistry to that in images. Results may improve with cleaner data. Similar to the Indigo set, the dataset was split into training and validation portions; 75% of the set was used to train with 25% reserved for validation.

Apart from the training and validation sets just described, two additional test sets were utilized in evaluating the performance of the method. The first is the dataset published with Valko & Johnson[7] (Valko dataset) consisting of 454 images of molecules cropped from literature. The Valko dataset is interesting because it contains complicated molecules with challenging features such as bridges, stereochemistry, and a variety of superatoms. The second dataset consists of a proprietary collection of image-SMILES pairs from 47 published articles and 5 patents. The molecules in the proprietary dataset are drug-like and some of the images contain small amounts of extraneous artifacts, e.g., surrounding text, compound labels, lines from enclosing table, etc. and was used to evaluate overall method effectiveness in examples extracted for use in real drug discovery projects.

With the focus of this work being on low quality/resolution images, rather than predicting high resolution images, we tested our method on downsampled versions of the Valko and proprietary datasets. During the segmentation phase each Valko dataset image was downsampled to 5-10% of its original size, and during the sequence prediction phase, images were downsampled to 10-22.5%, with similar scaling performed on the proprietary dataset. These scale ranges were chosen so that the resolution used during prediction approximately matched the (low) resolutions of the images used during training.

A list of 65 characters containing all the unique characters in the Indigo dataset was assembled. These characters served as the list of available characters that can be selected at each SMILES decoding step. Four of these characters are special tokens that are not part of SMILES notation but were necessary for successfully implementing the model. The special tokens indicate the beginning or end of a sequence, replace unknown characters, or pad sequences that were shorter than the maximum length (during training and testing 100 characters were generated for each input and any characters generated after the end-of-sequence token were ignored).

## Training

The segmentation model had 380,000 parameters and was trained on batches of 64 images (128x128 pixels in size). In our experiments, training converged after 650,000 steps and took 4 days to complete on a single GPU. The sequencing model had 46.3 million parameters and was trained on batches of 128 images (256x256 pixels in size). During training, images were randomly affine transformed, brightness scaled, and/or binarized. Augmenting the dataset while training using random transformations ensured that the model would not become too reliant on styles either generated by Indigo or seen in the patent images. In our experiments, training converged after 1 million training steps (26 days on 8 GPUs). Both models were constructed using TensorFlow[18] and were trained using the Adam optimizer[17] on NVIDIA Pascal GPUs.

## Results and Discussion

During training, metrics were tracked for performance on both the Indigo and USPTO validation datasets. We observed no apparent overfitting over the Indigo dataset during training but did experience some overfitting over USPTO data (Figure 6). Due to the large size of the Indigo dataset (52 million examples used during training) and the many rendering styles available in Indigo it is not surprising that the model did not experience any apparent overfitting on Indigo data. Conversely, the USPTO set is much smaller (1.27 million examples used during training) with each example sampled much more frequently (approximately 40 times more often), increasing the risk of overfitting.

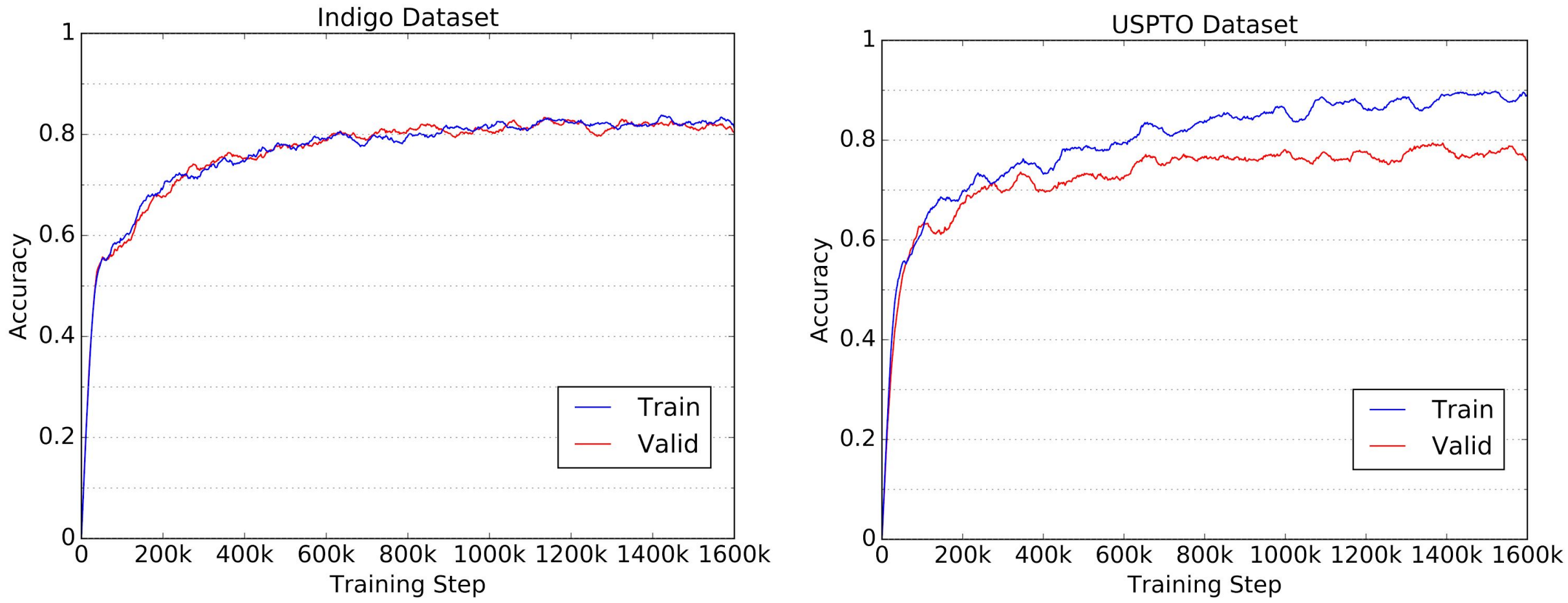

**Figure 6.** Training and validation curves for both the Indigo and USPTO datasets are here depicted.

After the models were trained, performance was measured on the Valko and proprietary test sets. The test sets were evaluated using the full segmentation and sequence generation pipeline described above, and accuracies for the validation and test sets are reported in Table 1. In order for a result to contribute to accuracy, it must be chemically identical to the ground truth, including stereochemistry. Any error results in the structure being deemed incorrect. We observed that despite the method requiring low resolution inputs, accuracy was generally high across the datasets. Additionally, accuracy for the validation sets and the proprietary set were all similar (77-83%) indicating that the training sets used in developing the method reasonably approximate data useful in actual drug discovery projects as represented by the proprietary test set.

**Table 1**

| Dataset | Accuracy |
|---|---|
| Indigo Validation | 82% |
| USPTO Validation | 77% |
| Valko Test Set | 41% |
| Proprietary Test Set | 83% |

On the Valko test set we observed an accuracy of 41% over the full, downsampled dataset, which is significantly lower than the accuracies observed in the other datasets. The decrease in performance is likely due to the higher rate of certain challenging features seen less frequently in the other datasets, including superatoms. Superatoms are the single largest contributor to prediction errors in the Valko dataset (21% of samples containing one or more incorrectly predicted superatoms). In our training sets, superatoms were only included in the USPTO dataset and were not generated as part of the Indigo dataset resulting in a low rate of inclusion during training (6.6% of total images seen contain some superatom, with most superatoms included at a rate of <<1%). An increased sampling rate of images containing superatoms will likely provide a significant accuracy improvement in this area.

In further exploring incorrectly generated superatoms we discovered, unsurprisingly, that larger or more uncommon superatoms were recognized with less success than smaller, more common types. For example, “Me” (methyl) is predicted correctly about half the time (other times being mistaken for a nitrogen or oxygen) while some larger superatoms are not predicted well at all (“$P^+(C_4H_9)_3$”, “$(H_2C)_5$”, and “$P^+$(n-Bu)$_3$” all predicted incorrectly). In some cases, however, large superatoms were recognized correctly, e.g., the single example of “n-$C_8H_{17}$” in the dataset is predicted correctly, and in Figure 7 we show an example structure with the “OTBS” (tert-butyldimethylsilyl ether) superatom predicted correctly despite aggressive downsampling and cluttering of characters.

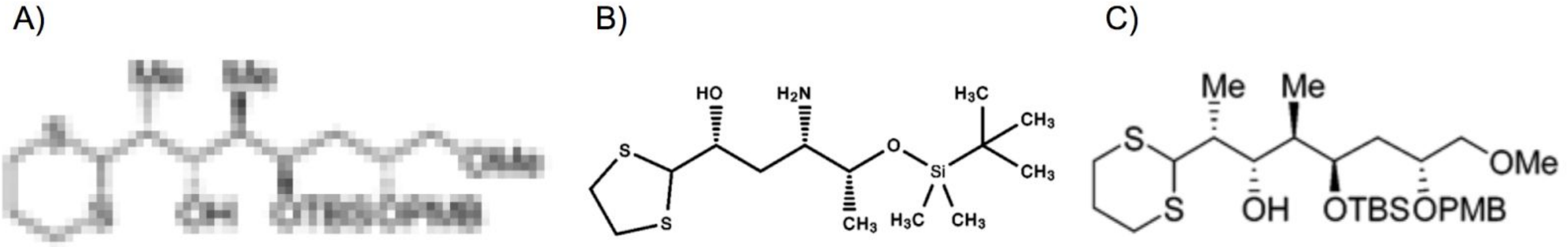

**Figure 7.** An example structure where a highly downsampled input (A) is used during prediction. The predicted structure (B) has a number of errors, likely due to this particular example being extremely low resolution, but the silyl ether is predicted correctly when compared against the ground truth (C).

Another interesting case in the dataset regarded the prediction of “$NEt_2$” (diethylamine) superatom. In Figure 8 three similar input images are shown, each containing the diethylamine functional group. In the results only one image had the functional group predicted correctly (the rightmost example in the figure) while the other two were incorrect, but interestingly the two incorrect examples were not incorrect in the same way. The middle example was predicted to contain an aniline while the leftmost example was predicted to contain an azide. This was despite the functional groups appearing nearly identical and occupying the same locality in the input images.

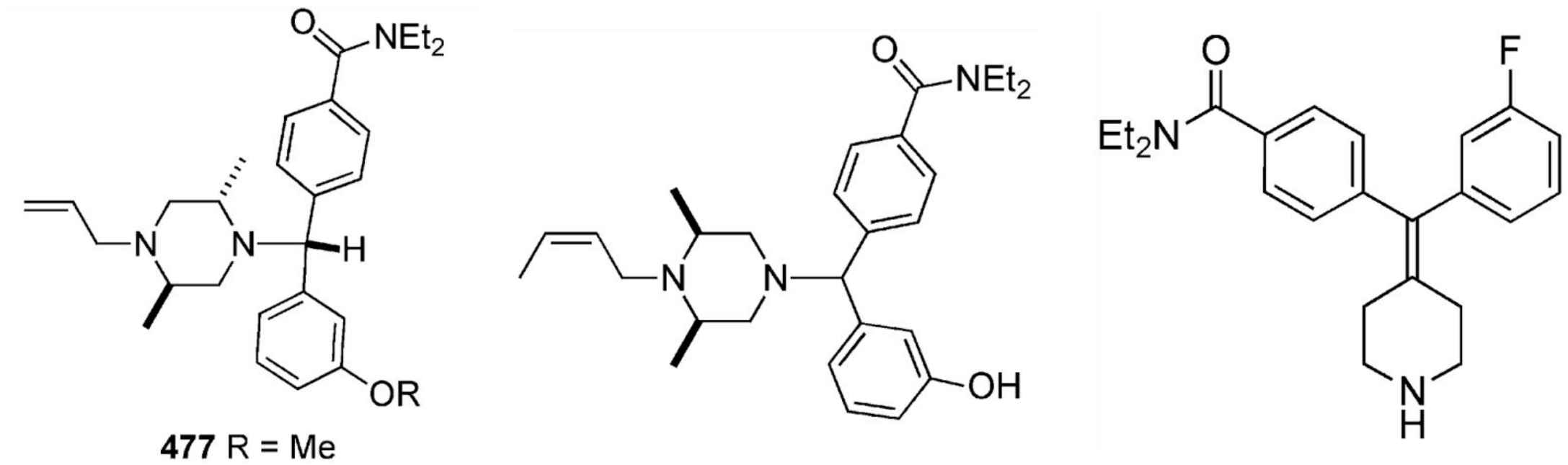

**477** R = Me
**478** R = H

N=[N+]=[N-]    NC1C=CC=CC=1    CCN(CC)

**Figure 8.** Three similar compounds containing the diethylamine superatom, only one of which was predicted correctly (rightmost). Below each image are the characters predicted in the SMILES for the diethylamine superatom.

In analyzing stereochemistry-related errors in the Valko dataset we observed that 60% of compounds with incorrectly predicted stereochemistry had explicitly assigned stereochemistry in both the ground truth and the predicted result, but the assignments in the predicted SMILES were incorrect. In other words, the model most often correctly predicted which atoms have explicit stereochemistry assigned, but occasionally assigned the wrong configuration (e.g., predicted R configuration when it should have been S). Intuitively, stereochemistry assignment is not a strictly local decision, i.e., observing a hash or a wedge is not sufficient information to make a configuration assignment, and more information about the neighboring atoms and connectedness is required for correct assignment. A possible explanation for the difficulty in learning stereochemistry from images is that our current model architecture may be insufficient in incorporating large enough context when computing certain features.

Some images failed to produce a valid structure (either output SMILES was not valid or output confidence was <1%). Common issues that resulted in a structure failing or otherwise being severely incorrect included structures that were too large, macrocycles with large rings that were cleaved during prediction, structures with many superatoms or stereocenters, or images where downscaling was too aggressive and resolution too low for adequate recognition. The Valko set also contains images with multiple structures or that are inverted (white structures on black background), neither of which were supported in our validation scheme and are reported as incorrect.

Across both test sets we observed low error rates due to segmentation and predicted masks appeared quite clean and were generated at reasonably high resolution (see Figure 2 in the method section above for an example). Only 3.3% of the Valko dataset and 6.6% proprietary images failed to segment properly.

To further analyze performance over the validation and test sets, we explored distributions for several metrics. In Figure 9 we report distributions of correct and incorrect examples for molecular weight, number of heavy atoms contained in the molecules, number of characters in the ground truth SMILES, and the types of heavy atoms contained in the molecules. In exploring molecular weight and heavy atoms of both correct and incorrect molecules in the USPTO validation set we observed that the model slightly favors smaller molecules. Predicting more errors on larger molecules was not surprising considering large molecules have longer SMILES strings and necessitates the model to compress more information during encoding and predict more characters during decoding. It was surprising, however, that the difference between correct and incorrect distributions is not more pronounced. Our expectation was that larger molecules would be significantly more challenging to predict and that the model would heavily favor smaller molecules. Incorrect SMILES tend to shift toward heavier or larger molecules, but correctness cannot be adequately attributed to either metric. Predicting well on large molecules is encouraging, and suggests that the model may be easily extended to molecules larger than 100 characters in SMILES length. In exploring the types of heavy atoms seen in both the correct and incorrect examples, once again, both distributions appeared similar. Particularly interesting are the atoms that appear much more rarely in SMILES, e.g., Na, Sn, W. In predicting rarer atoms, the network performed surprisingly well on some (Na, Ar) but not well at all on others (U, V). Further work is needed to explore the distribution of rare atoms across the full dataset and ensure that all atom types are sampled sufficiently during training.

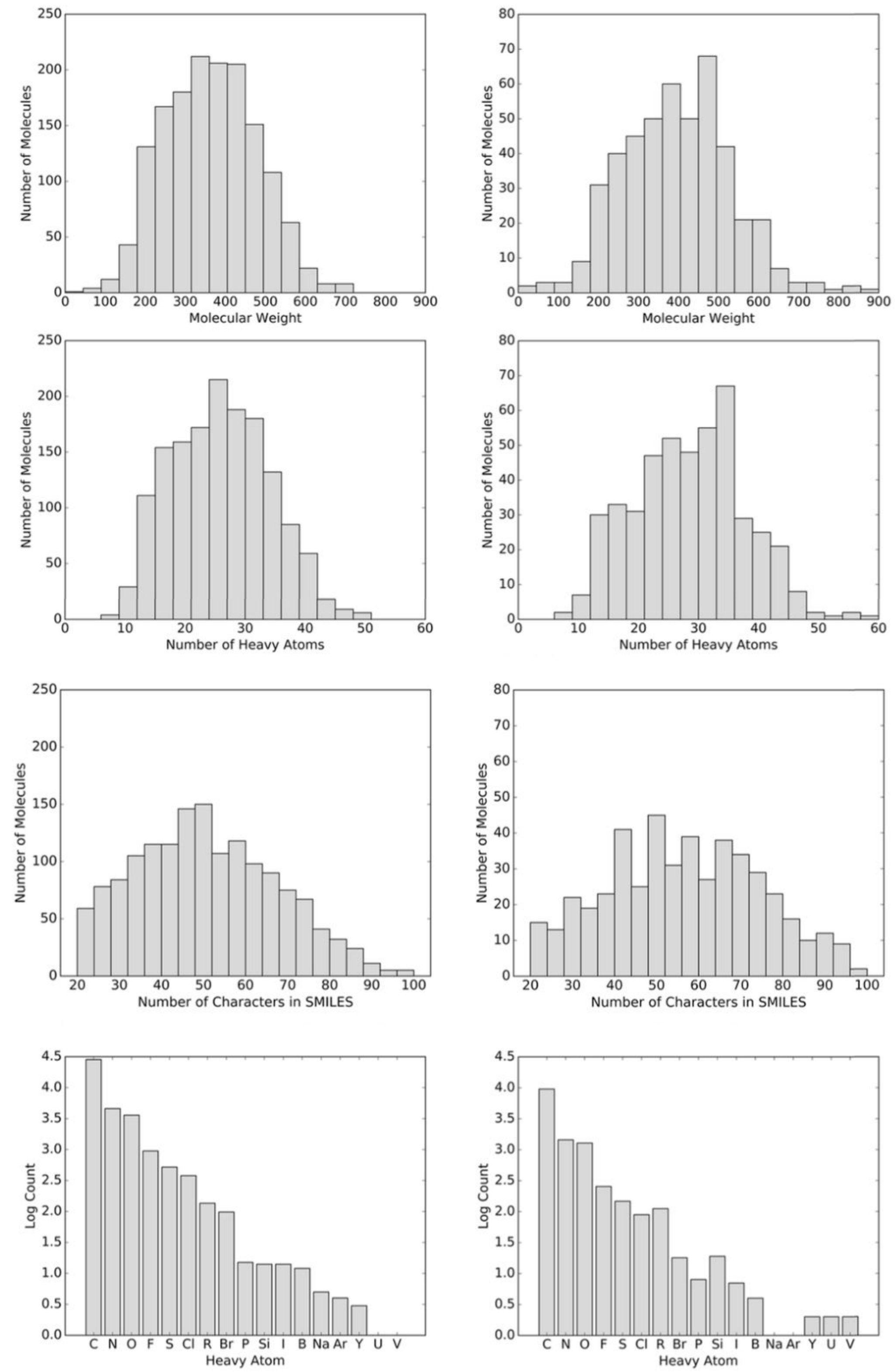

**Figure 9.** Distributions of metrics for both correct (left) and incorrect (right) predictions from the USPTO validation set. Statistics for incorrect examples were taken from the reference SMILES.

Similar to the analysis on the USPTO dataset reported above, we explored distributions of simple molecular properties for SMILES predicted correctly in the Valko dataset (Figure 10). Interestingly, the distributions all appear to be more narrow than in the USPTO dataset and the SMILES strings are longer. It is worth noting that the Valko dataset is quite small and a larger dataset containing a broader distribution of molecules would be interesting for the community to benchmark against, but is left for future work.

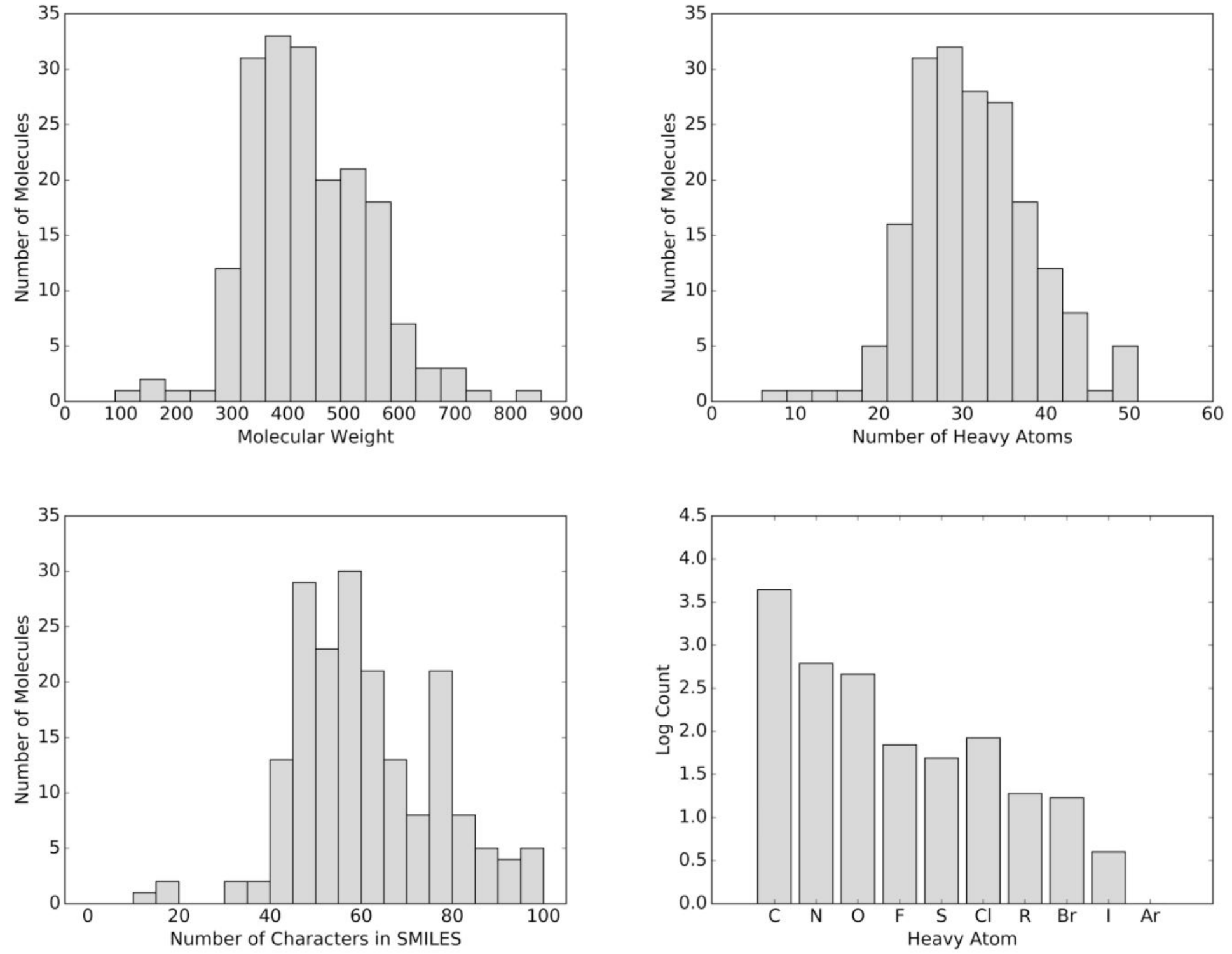

**Figure 10.** Distributions of metrics for correct predictions from the Valko et al. test set.

In reviewing structures that were predicted correctly, we observe that the methods described in this work show promise in their ability to predict valid and correct SMILES for low resolution images. We showcase a few examples in Figure 11. These examples contain a variety of atom types, some examples of stereochemistry and superatoms, and are not trivial in size. Further progress may require developing methods which eliminate the restriction of downsampling all inputs by supporting high resolution data when available, and supporting structures larger than 100 characters in length.

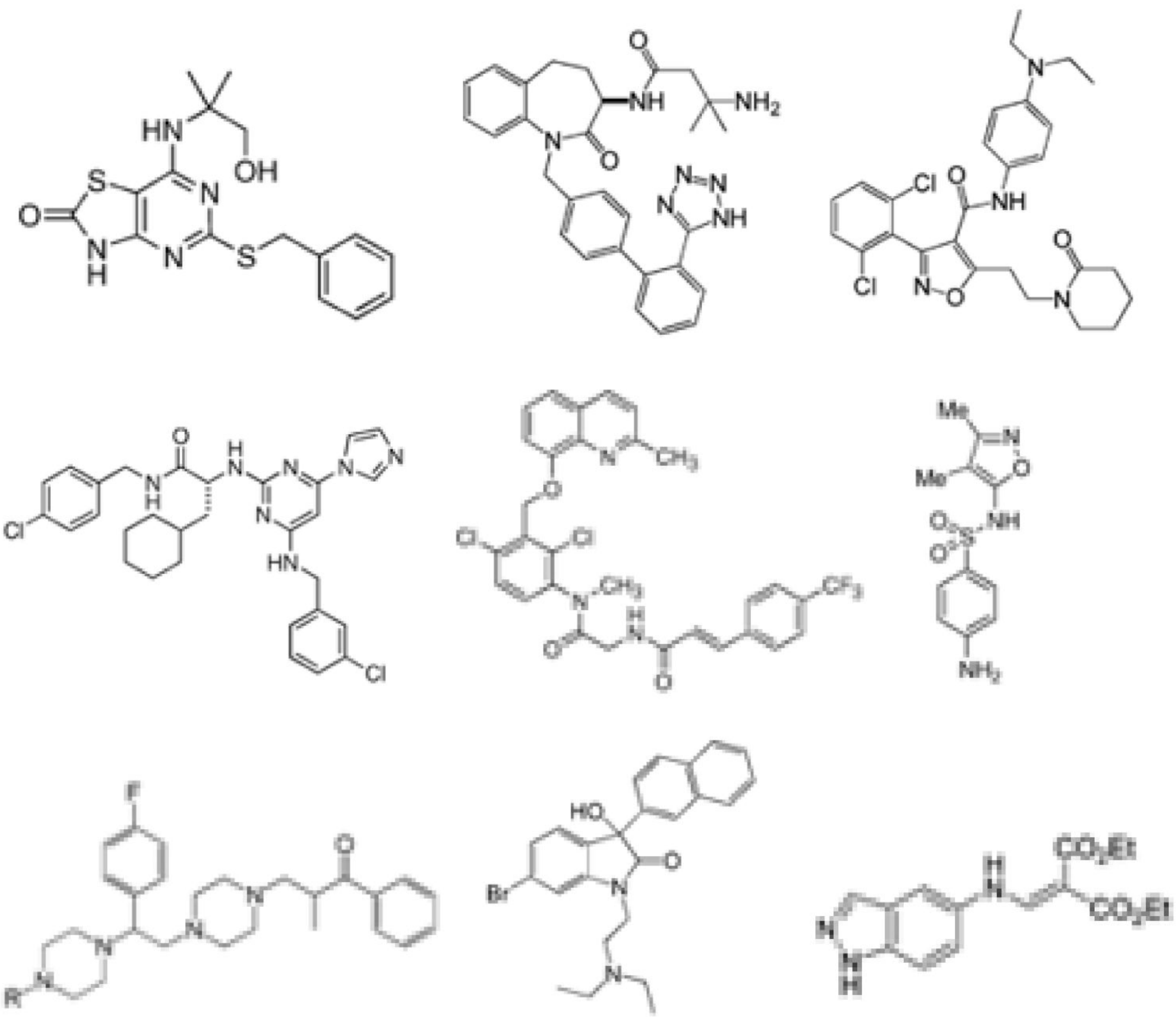

**Figure 11.** Examples of low resolution structures predicted correctly.

## Conclusion

In this work we presented deep learning solutions to both extract structures from documents and predict SMILES for structure images. The method does not rely on handcrafted features or rules and operates directly on raw pixels enabling the method to learn from and predict images of virtually any style. Using datasets containing molecule images cropped from journal articles and patents we showed that deep learning can learn to predict images of molecules from literature at reasonably high accuracy. The method herein was trained exclusively on low resolution data, and thus only supported prediction over low resolution input. Training over high resolution images as well may greatly improve results, particularly when high resolution inputs are available. All images used in the reported results were highly downsampled, demonstrating the ability to predict low resolution images of chemical structures using an automated method, which was not previously possible.

We anticipate the use of chemical structure extraction algorithms, such as those described herein as well as future generalizations and improvements, may greatly accelerate drug discovery efforts in many ways. Most immediately, such algorithms may help to greatly accelerate curation of published journal article and patent data to facilitate routine QSAR/QSPR modelling work. However, given the very high rate at which data is being introduced into the public academic and patent literature, expeditious and efficient curation of public data may in the future become a chief bottleneck in the construction of maximally optimal global ADMET property prediction models for drug discovery. Steps toward fully automating data curation may enable drug discovery projects to more routinely utilize all relevant available data for ADMET property prediction at all moments in time in the progression of the project. Given the widespread recognition of the dependence of ADMET property prediction on data set size and cleanliness, we anticipate such technologies should broadly improve the quality of drug discovery ADMET property modeling in the future.

## Acknowledgements

The authors would like to thank Hercules Silverstein for documenting and refactoring code, Shawn Watts and Ken Dyall for useful discussions, and the Schrödinger Machine Learning and Data Teams for useful discussions and feedback.